\documentclass[11pt]{article}

\usepackage[preprint]{latex/acl}

\usepackage{times}
\usepackage{latexsym}

\usepackage[T1]{fontenc}

\usepackage[utf8]{inputenc}

\usepackage{microtype}

\usepackage{inconsolata}

\usepackage{graphicx}

\usepackage[inline]{enumitem}
\usepackage{amssymb}
\usepackage{amsmath}
\usepackage[british]{babel}
\usepackage[babel=true]{csquotes}
\usepackage{listings}
\usepackage{xurl}
\usepackage{booktabs}
\usepackage{float}
\usepackage[table]{xcolor}
\usepackage{color}
\usepackage{graphicx}
\usepackage{multirow}
\usepackage{tabularx}
\newcolumntype{C}{>{\centering\arraybackslash}X}
\newcolumntype{L}{>{\arraybackslash}X}
\newcolumntype{x}[1]{>{\centering\arraybackslash\hspace{0pt}}p{#1}}

\usepackage[acronym]{glossaries}
\glsdisablehyper

\newcommand{\gradientcell}[6]{
    \ifdimcomp{#1pt}{>}{#3 pt}{\cellcolor{#5!100.0!#4!#6}#1}{
    \ifdimcomp{#1pt}{<}{#2 pt}{\cellcolor{#5!0.0!#4!#6}#1}{
         \pgfmathparse{int(round(100*(#1/(#3-#2))-(#2 *(100/(#3-#2)))))}
        \xdef\tempa{\pgfmathresult}
        \cellcolor{#5!\tempa!#4!#6}\hspace*{-3.5pt}#1\hspace*{-4.5pt}
    }}
 }
\usepackage{pgf} 
\usepackage{colortbl}
\usepackage[multiple]{footmisc}

\newacronym{llm}{LLM}{Large Language Model}
\newacronym{ai}{AI}{Artificial Intelligence}
\newacronym{toefl}{TOEFL}{Test of English as a Foreign Language}
\newacronym{pdf}{PDF}{Probability Density Function}
\newacronym{std}{SD}{standard deviation}
\newacronym{nb}{NB}{na\"{i}ve Bayes}
\newacronym{fpr}{FPR}{False Positive Rate}
\newacronym{fnr}{FNR}{False Negative Rate}
\newacronym{acc}{Acc}{accuracy}
\newacronym{unk}{Unk}{unknown token rate}
\newacronym{rda}{RDA}{random data augmentation}

\newcommand{\entropy}{entropy}
\newcommand{\qi}{\textbf{Q1}}
\newcommand{\qii}{\textbf{Q2}}
\newcommand{\qiii}{\textbf{Q3}}

\let\origfootnote\footnote
\renewcommand{\footnote}[1]{\kern.1em\origfootnote{#1}}
\newcommand{\punctfootnote}[1]{\kern-.1em\origfootnote{#1}}

\newcommand{\SYNnat}{\(\textbf{\textsc{SYNv9}}_\textbf{\textsc{auth}}\)}
\newcommand{\SYNGPT}{\(\textbf{\textsc{SYNv9}}_\textbf{\textsc{GPT4o}}\)}
\newcommand{\SYNtrain}{\(\textbf{\textsc{SYNv9}}^\textbf{\textsc{train}}\)}
\newcommand{\SYNval}{\(\textbf{\textsc{SYNv9}}^\textbf{\textsc{val}}\)}
\newcommand{\SYNmini}{\(\textbf{\textsc{SYNv9}}_\textbf{\textsc{4oMini}}^\textbf{\textsc{val}}\)}
\newcommand{\SYNllama}{\(\textbf{\textsc{SYNv9}}_\textbf{\textsc{Llama}}^\textbf{\textsc{val}}\)}
\newcommand{\Wikinat}{\(\textbf{\textsc{Wiki}}_\textbf{\textsc{auth}}\)}
\newcommand{\WikiGPT}{\(\textbf{\textsc{Wiki}}_\textbf{\textsc{GPT4o}}\)}
\newcommand{\Wiki}{\(\textbf{\textsc{Wiki}}\)}
\newcommand{\Newsnat}{\(\textbf{\textsc{News}}_\textbf{\textsc{auth}}\)}
\newcommand{\NewsGPT}{\(\textbf{\textsc{News}}_\textbf{\textsc{GPT4o}}\)}
\newcommand{\News}{\textbf{\textsc{News}}}
\newcommand{\NonNative}{\textbf{\textsc{Non\-Na\-tive}}}
\newcommand{\NonNativeC}{\textbf{\textsc{Non\-Na\-tive\-C1}}}
\newcommand{\NatYouth}{\textbf{\textsc{Nat\-Youth}}}
\newcommand{\NatAdv}{\textbf{\textsc{Nat\-Adv}}}
\newcommand{\AbsOld}{\textbf{\textsc{Abs\-2020}}}
\newcommand{\AbsNew}{\textbf{\textsc{Abs\-New}}}

\title{Different Time, Different Language: Revisiting the Bias Against Non-Native Speakers in GPT Detectors}

\author{Adnan Al Ali\textsuperscript{1} \and Jindřich Helcl\textsuperscript{2} \and Jindřich Libovický\textsuperscript{1} \\
\textsuperscript{1} Charles University, Faculty of Mathematics and Physics \\
\textsuperscript{2} University of Oslo, Language Technology Group \\
  \texttt{alali@ufal.mff.cuni.cz} \\
  }

\begin{document}
\maketitle


\begin{abstract}
    \acrshort{llm}-based assistants have been widely popularised after the release of ChatGPT. Concerns have been raised about their misuse in academia, given the difficulty of distinguishing between human-written and generated text. To combat this, 
    automated techniques have been developed 
    and shown to be effective, to some extent. However, prior work suggests that these methods often falsely flag essays from non-native speakers as generated, due to their low perplexity extracted from an \acrshort{llm}, which is supposedly a key feature of the detectors. We revisit these statements two years later, specifically in the Czech language setting. We show that the perplexity of texts from non-native speakers of Czech is \emph{not} lower than that of native speakers. We further examine detectors from three separate families and find no systematic bias against non-native speakers. Finally, we demonstrate that contemporary detectors operate effectively without relying on perplexity.
\end{abstract}

\section{Introduction}

Following the release of \acrshort{llm}-based assistants -- most notably ChatGPT, which was based on GPT-3 \citep{gpt3} and upgraded to GPT-4 \citep{gpt4} and later versions
-- and their subsequent growth in popularity, concerns have emerged about the possible misuse of the service, particularly for plagiarism. This concern was largely raised in academic contexts \citep{gpt-education}.
\begin{figure}[t]
  \includegraphics[width=\columnwidth]{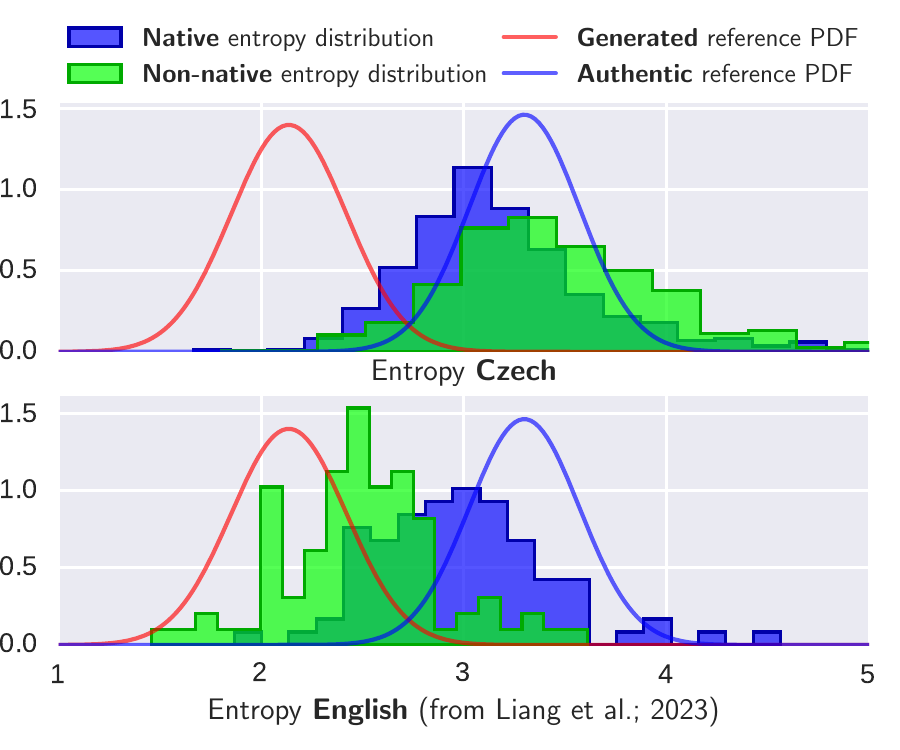}
  \caption{Distribution density of the \entropy{} extracted from an \acrshort{llm} for essays from \colorbox[RGB]{155,155,255}{native} vs. \colorbox[RGB]{155,255,155}{non-native} speakers of Czech (top) and English (bottom). Unlike the English essays \citep{detectors_biased}, we find that essays written by non-native speakers of Czech have higher entropies on average than those of their native peers. The reference \acrshort{pdf} was computed on a Czech corpus.
  }
  \label{fig:motivation}
\end{figure}

Given the natural-sounding text generation, the distinction between human-written and generated text is challenging for humans \citep{ippolito-etal-2020-automatic-human-det, czech-generated-detection}.
In contrast, machine-learning methods proved to be accurate to some extent \citep{detection-survey}. However, according to \citet{detectors_biased}, some of these methods are perplexity-based\footnote{Perplexity in this context is measured with respect to an \acrshort{llm}; see Section~\ref{sec:entropy-perplex} for definitions.}
and tend to be biased against non-native speakers of English, whose texts often have lower perplexities.
As a result, the texts from non-native speakers might be
falsely flagged as \acrshort{ai}-generated.

In this article, we follow up on the work of \citet{detectors_biased} in the Czech-speaking context and aim to answer three fundamental questions:
\begin{enumerate}[label=\textbf{Q\arabic*},noitemsep]
    \item \label{q1} Is the perplexity of the texts from non-native speakers of Czech lower than that of the texts of their native peers?
    \item \label{q2} Is there a bias against non-native speakers in Czech generated text detectors?
    \item \label{q3} Is it possible to create generated text detectors without -- explicitly or implicitly -- relying on perplexity?
\end{enumerate}


To answer \qi, we use an \entropy{}-based
analyser. Note that \entropy{} and perplexity have a monotonic exponential relationship. We measure the \entropy{} distributions in all the inspected domains and compare the texts from non-native speakers with those of their native peers, as well as with texts from other domains.

For \qii, we examine a set of generated text detectors from the commonly used categories:
\begin{enumerate*}[label=(\arabic*)]
    \item classical machine learning model using a bag-of-words text representation,
    \item fine-tuned pre-trained RoBERTa-like model, and 
    \item closed-source commercial detector.
\end{enumerate*}
We evaluate these models across multiple domains to assess their overall quality and performance on texts from native and non-native speakers.

Finally, to answer \qiii, we inspect the correlations within one class (human-written or generated) between the outputs of the \entropy{}-based analyser and the detectors, to assess whether they implicitly work with some internal representation of the \entropy{} (or possibly explicitly in the case of the closed-source detector). We further examine how these correlations change across domains.

Our research reveals that the answers to the three proposed questions differ considerably from those presented in the seminal work of \citet{detectors_biased}, which has been reported in mainstream news outlets.
The language setting proves to be a significant factor in considering the bias against non-native speakers.

The paper is structured as follows: Section~\ref{sec:related-work} discusses the related previous work. Section~\ref{sec:datasets} describes the creation of the datasets used in this work and their significance. In Section~\ref{sec:entropy-perplex}, we define the terms perplexity and entropy and use entropy analysis to address question~\qi{}. In Section~\ref{sec:detectors}, we create and evaluate \acrshort{llm} detectors, addressing question~\qii{}. Section~\ref{sec:correlation} discusses how the entropy impacts the predictions of the detectors, addressing question~\qiii{}. Finally, we conclude our findings in Section~\ref{sec:conclusion}.



\section{Related Work}\label{sec:related-work}

The concern of using \acrshort{llm}-generated text for academic misconduct has been raised since the release of ChatGPT. \citet{gpt-education}\footnote{Preprint published in 2022.} provided an early examination of the capabilities of ChatGPT in academic settings. The study underscored a concern that \acrshort{llm}s may pose a threat to academic integrity, especially in online examinations. The authors state \acrshort{llm} detectors as one of the prominent countermeasures to combat plagiarism.

\subsection{Generated Text Detection}

To our knowledge, no studies have been published on the creation/training of \acrshort{llm} detectors in Czech. Nonetheless, several multilingual evaluation benchmarks for \acrshort{llm} detectors contain Czech samples, such as \emph{MULTITuDE} \citep{macko-etal-2023-multitude}, which is based on a dataset of news articles \citep{varab-schluter-2021-massivesumm} and complemented with \acrshort{llm}-generated counterparts. The authors show that multilingual detectors fine-tuned on English, Spanish, and Russian samples can be zero-shot-transferred to Czech and maintain an \(F_1\) score greater than 0.85. However, this dataset is limited to the news domain, which is a significant limitation, as detectors tend to be sensitive to domain changes (see below).


Various studies have been conducted on the automatic detection of generated text in English. \citet{detection-survey} provide a comprehensive survey on the problem. Below, we list three prominent approaches; however, this list is not exhaustive. 

\emph{Classical machine learning methods} using bag-of-words features in combination with SVMs, random forests, and logistic regression, among others, achieve 
performance comparable to more complex methods \citep{openai-old-detector, svm-detector} and serve as a solid baseline.

\emph{Logit-based methods} use the raw outputs of a
reference \acrshort{llm}. \citet{openai-old-detector} showed that the log-likelihood of a text under a model (the opposite value of \entropy{}) is a useful feature but not satisfactory by itself, as this value differs across domains \citep{howkgpt}. More complex logit-based methods have been successfully used for semi-automatic \citep{gltr} and automatic \citep{log-rank,detectgpt} detection.

Fine-tuning a \emph{pre-trained language model}, such as BERT \citep{bert}, has been a common approach \citep{openai-old-detector,fagni2021tweepfake,gptsentinel}. However, such models have been shown to lack robustness when new domains or misspellings are introduced \citep{bert-not-robust}, which is a key limitation.



\subsection{Bias in GPT Detectors}

\setlength{\tabcolsep}{4pt}
\begin{table*}[t]
    \centering\footnotesize
    \begin{tabularx}{0.90\linewidth}{c L c c c}
        \toprule
        \textbf{Dataset} & \textbf{Description} & \textbf{\#samples} & \textbf{Avg. \#tokens} & \textbf{\acrshort{llm}?} \\
        \midrule
        \SYNtrain{} & Czech National Corpus + GPT-4o complement (train) & 4766 & 511.57 & Mix \\

        \SYNval{} & Czech National Corpus + GPT-4o complement (val) & 598 & 511.45 & Mix \\
        
        \SYNmini{} & Czech National Corpus 4o-mini complement & 287 & 512.00 & Yes  \\
        
        \SYNllama{} & Czech National Corpus Llama complement & 301 & 510.55 & Yes  \\
        
        \Wiki{} & Wikipedia crawl + GPT-4o complement & 422 & 512.00 & Mix \\
        
        \News{} & News crawl + GPT-4o complement & 762 & 511.94 & Mix \\
        
        \NonNative{} & Essays from non-native speakers & 450 & 342.16 & No \\
        
        \NonNativeC{} & Essays from proficient non-native speakers & 29 & 512.00 & No \\
        
        \NatYouth{} & Essays from native speakers (children) & 450 & 331.51 & No \\
        
        \NatAdv{} & Essays from native speakers (age 16--18) & 29 & 496.59 & No \\
        
        \AbsOld{} & Pre-GPT theses abstracts & 1655 & 283.29 & No \\
        
        \AbsNew{} & Post-GPT theses abstracts & 2050 & 298.51 & Unk \\
        \bottomrule
    \end{tabularx}
    \caption{Overview of the datasets, their sample count, average number of \texttt{unitok} tokens per sample (after truncation to max. 512 tokens), and their \acrshort{llm}-generated status. While it is unclear whether (and to what extent) \AbsNew{} is generated, the metrics in this article assume it is human-written for simplicity.}
    \label{tab:data-overview}
\end{table*}

\citet{detectors_biased} examined how detectors of \acrshort{llm} work on text written by non-native speakers and found that the detectors systematically flag the texts written by them as generated. The authors evaluated seven \enquote{widely used} generated text detectors on two groups of documents:
\begin{enumerate*}[label=(\arabic*)]
    \item \acrfull{toefl} essays written by Chinese students (91 documents), and
    \item Hewlett Foundation’s ASAP dataset \citep{asap-aes}, containing US eight-graders' essays (88 documents).
\end{enumerate*}

The study found that, in most cases, the evaluated detectors correctly labelled the essays from US students as human-written, with the mean 
\acrfull{fpr} being 5.1\%. In contrast, the detectors often misclassified the \acrshort{toefl} essays written by Chinese students as GPT-generated, with the mean \acrshort{fpr} of 61.3\%. Furthermore, all seven detectors unanimously flagged 19.8\% of the \acrshort{toefl} essays as \acrshort{ai}-generated. Those essays have been shown to have low perplexities.

The authors further proceed to present a claim that \emph{\enquote{most GPT detectors use text perplexity to detect \acrshort{ai}-generated text}}. In our study, we follow up on the presented claims and test whether the texts from non-native speakers of Czech have smaller perplexities and whether we can create a classifier that does not rely on perplexity. Our findings in the Czech setting differ significantly from those in the prior study.


\section{Datasets}\label{sec:datasets}


Training and evaluation of detectors of \acrshort{llm}s requires carefully curated datasets of clear origin (human or \acrshort{llm}). For training, a reasonably large corpus (comprising millions of tokens) of high-quality text is required (both human-written and generated). Evaluation data must encompass diverse domains extending beyond the training data. Importantly for our purposes, the evaluation data must also contain texts from non-native speakers and comparable texts from native speakers. Table~\ref{tab:data-overview} contains the overview of our datasets.





\subsection{Conteporary Czech Corpus}

For training, we exclusively use SYNv9 \citep{synv9}, which is the most comprehensive collection of contemporary (synchronic) Czech corpora consisting of news/magazines (predominantly), non-fiction, and fiction domains. 
We randomly sampled 7460 texts published between the years 2009 and 2019.
We truncated the texts to 2000 pre-annotated tokens. This dataset of authentic documents is referred to as \SYNnat{}.

We complement the authentic data with an \acrshort{llm}-generated counterpart.
To match the structure and vocabulary of the authentic corpus, we used the prompt that contained a short sample of the texts from \SYNnat{}, resulting in one generation prompt for each text
(see Appendix~\ref{app:dataset} for details).

We produced generated samples using various \acrshort{llm}s. As the primary source of generated text, we use GPT-4o\footnote{Version \texttt{gpt-4o-2024-05-13}.} \citep{gpt-4o}, with the temperature 0.7 and number of tokens limited to 1024. We generated the data, discarded the files smaller than 2~kB, and split them into two subsets: \(\textbf{\textsc{SYNv9}}_\textbf{\textsc{GPT4o}}^\textbf{\textsc{train}}\) and \(\textbf{\textsc{SYNv9}}_\textbf{\textsc{GPT4o}}^\textbf{\textsc{val}}\). For training, we paired the \(\textbf{\textsc{SYNv9}}_\textbf{\textsc{GPT4o}}^\textbf{\textsc{train}}\) samples with the authentic samples that were used to generate them, resulting in the training set 
\SYNtrain{}. Analogously, we created \SYNval{}.

To include more models for validation, we created \SYNmini{} using GPT-4o-mini\footnote{Version \texttt{gpt-4o-mini-2024-07-18}; \url{https://openai.com/index/gpt-4o-mini-advancing-cost-efficient-intelligence}} and \SYNllama{} using Llama 3.1 405B\footnote{Ollama model \texttt{llama3.1:405b-instruct-q5\_K\_S}} \citep{llama3} analogously.

\subsection{Wikipedia and Online News Crawl}

To include more domains, we added Wikipedia and online news articles for evaluation. The \Wikinat{} dataset was created by crawling Wikipedia using \texttt{pywikibot}.\punctfootnote{\url{https://www.mediawiki.org/wiki/Manual:Pywikibot}} Articles were chosen randomly, retrieved and parsed using \texttt{mwparserfromhell}.\punctfootnote{\url{https://github.com/earwig/mwparserfromhell}} Articles that contained fewer than 1000 space characters were discarded. The generated complement was created using GPT-4o, prompted to write a Wikipedia article on a given topic from \Wikinat{}. After filtering, 211 articles were created (\WikiGPT{}) and paired with their authentic counterparts, creating \Wiki{}.

We sampled the \Newsnat{} dataset randomly from the web news crawl 2021 \citep{kocmi-EtAl:2022:WMT}.\punctfootnote{Although the news domain is contained in \(\textbf{\textsc{SYNv9}}\), the structure of online news articles may likely be different from the printed ones.}
Again, we generated a complement using GPT-4o with an analogous prompt, creating 381 generated articles (\NewsGPT{}) and paired them with their counterparts, resulting in \News{}.

\subsection{Non-Native and Native Youth Works}

As a crucial dataset for determining the performance of our classifiers on text by non-native speakers, we utilised the AKCES~3 corpus \citep{akces3}, a corpus of essays written by non-native students of the Czech language. To filter out texts with too frequent mistakes, we only included speakers who had studied Czech for at least 24 months at the time of writing. The dataset is referred to as \NonNative{}.

In an effort to better reproduce the work of \citet{detectors_biased}, we created \NonNativeC{}, a dataset of advanced non-native speakers, sourced from an extended version of AKCES~3 \citep{gecc}, from speakers with language proficiency labelled as proficient.\punctfootnote{C1 or C2 under the CEFR.} Finally, we filtered the 118 samples, which we found to still contain frequent errors, to be at least 2~kB in size, yielding 29 samples, predominantly from Slavic authors. We state the distributions of the L1 languages of the non-native datasets in Appendix~\ref{app:l1-langs}.

To roughly match the domain of \NonNative{}, we utilised the AKCES~1 corpus \citep{akces1}, a collection of essays written by native Czech speakers at primary (from 5th grade) and secondary schools. Furthermore, we attempted to match the distribution of the file size of the native texts to the \NonNative{} dataset by selecting the most similar (in file size) text from AKCES~1 for each text in the \NonNative{} dataset. We denote the resulting subset of AKCES~1 by \NatYouth{}.

To match the advanced non-native texts from \NonNativeC{}, we randomly selected 29~texts from AKCES~1 with the age category labelled as \enquote{over 15~years} -- i.e. 16--18 years. We denote this dataset by \NatAdv{}.

\subsection{Academic Abstracts}

To evaluate the models on academic texts, we created a corpus of theses abstracts, crawled from the Charles University Digital Repository,\punctfootnote{\url{https://dspace.cuni.cz}; the largest database of theses in Czech.}
published between 2020 and 2021 -- i.e., before the introduction of ChatGPT. We denote this dataset by \AbsOld{}.

For comparison, we also included the abstracts written between 2023 and 2025 -- after ChatGPT became widely popular -- yielding \AbsNew{}.

\section{Entropy and Perplexity}\label{sec:entropy-perplex}

To address question \qi{} -- i.e. whether non-native speakers of Czech tend to produce texts that LLMs rate with smaller perplexity compared to their native peers -- we redefine the task using entropy and analyse the datasets. In this context, perplexity is defined as the \textit{\enquote{exponential average negative log-likelihood of a token sequence given a specific language model}} \citep{detectors-not-biased}.
\setlength\abovedisplayskip{0.4em}
\setlength\belowdisplayskip{0.4em}




Omitting the exponentiation, we can define \entropy{} as the per-token average negative log-likelihood of a document given an \acrshort{llm}:
\begin{equation}\label{eq:entropy}
    -\frac{1}{N} \sum_{i=1}^N \log P(d_i \mid d_1, \dots, d_{i-1})
\end{equation}
where \((d_1, \dots, d_N)=\mathbf{d}\) is a token sequence. 

We chose to work with \entropy{} rather than perplexity, as it roughly follows a Gaussian distribution (as shown in Figure~\ref{fig:syntrain-entropy}). For our purposes, the two metrics are interchangeable, as they have a monotonically increasing relationship, preserving the ordering.

\subsection{Entropy Analysis}
For our reference model, we chose Llama 3.2 1B base \citep{llama3}, unlike previous work \citep{detectors_biased,detectors-not-biased}, which used GPT-2 \citep{gpt2}. This is because GPT-2 performs poorly on Czech \citep{CzeGPT} and is somewhat outdated.\punctfootnote{Another option would be GPT-OSS \citep{openai2025gptoss120bgptoss20bmodel}. However, this model is not published in its base version, possibly making the entropy calculation less reliable.} Nonetheless, we found that our Llama-based entropy analyser produces comparable results to those reported by \citet{detectors_biased} when applied to their dataset (see Table~\ref{tab:ent-mean-std} bottom and Figure~\ref{fig:motivation}).



The number of samples for each dataset was clipped to 1000, and the subset was selected at random.
We truncated each sample to 512 tokens and disregarded the predictions on the first 50 tokens to provide sufficient context, for better stability. Let \(M\) denote the length of the truncated document, then our modified entropy formula is as follows:
\begin{equation}\label{eq:our-entropy}
-\frac{1}{M - 50} \sum_{i=51}^M \log P(d_i \mid d_1, \dots, d_{i-1})
\end{equation}
We display the distribution density together with the fitted Gaussian \acrshort{pdf}s for \SYNtrain{} in Figure~\ref{fig:syntrain-entropy}
and the distributions for the remaining datasets in Appendix~\ref{app:entropy-dist}.
In most cases, the generated documents have smaller entropies than authentic, although some overlap is present.

\begin{figure}
    \centering
    \includegraphics[width=\columnwidth]{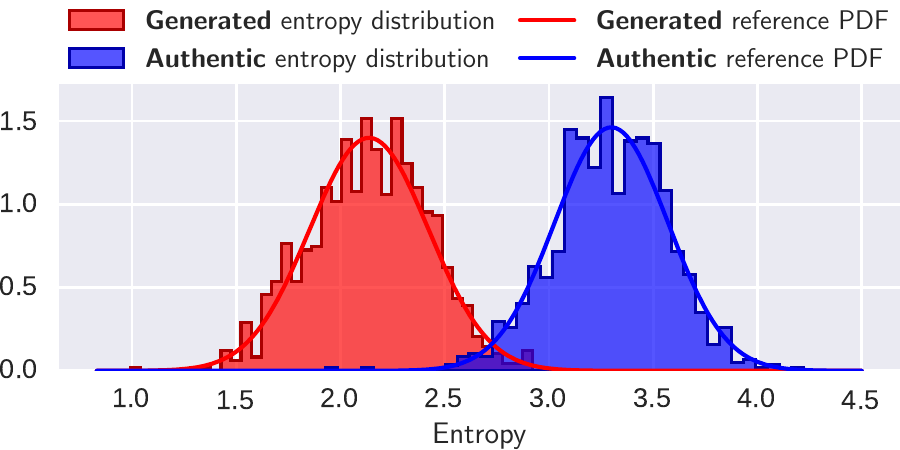}
    \caption{Distribution density of the \entropy{} for \colorbox[RGB]{255,155,155}{generated} and \colorbox[RGB]{155,155,255}{authentic} samples from \SYNtrain{}, together with their fitted Gaussian \acrshort{pdf}.}
    \label{fig:syntrain-entropy}
\end{figure}

\newcommand{\Ent}[1]{\gradientcell{#1}{1.66}{3.45}{cyan}{yellow}{70}}
\newcommand{\EntSD}[1]{\gradientcell{#1}{0.18}{0.58}{cyan}{yellow}{35}}

\setlength{\tabcolsep}{7pt}
\begin{table}[ht]
    \centering\footnotesize
    \begin{tabularx}{\columnwidth}{ C c c c c }
        \toprule
        \multirow{2}{*}[-0.3em]{\textbf{Dataset}} & \multicolumn{2}{c}{\textbf{Generated}} &  \multicolumn{2}{c}{\textbf{Natural}} \\
        \cmidrule{2-3}\cmidrule{4-5}
         & \textbf{Mean} & \textbf{\acrshort{std}} & \textbf{Mean} & \textbf{\acrshort{std}} \\
         \midrule
        \SYNtrain{} & \Ent{2.14} & \EntSD{0.29} & \Ent{3.30} & \EntSD{0.27} \\
        
        \SYNval{} & \Ent{2.13} & \EntSD{0.29} & \Ent{3.30} & \EntSD{0.27} \\
        
        \SYNmini{} & \Ent{2.09} & \EntSD{0.26} & -- & -- \\
        
        \SYNllama{} & \Ent{1.97} & \EntSD{0.36} & -- & -- \\
        
        \Wiki{} & \Ent{1.67} & \EntSD{0.20} & \Ent{2.4} & \EntSD{0.30} \\
        
        \News{} & \Ent{1.89} & \EntSD{0.18} & \Ent{2.82} & \EntSD{0.36} \\
        
        \NonNative{} & -- & -- & \Ent{3.48} & \EntSD{0.57}  \\
        
        \NonNativeC{} & -- & -- & \Ent{2.97} & \EntSD{0.36}  \\
        
        \NatYouth{} & -- & -- & \Ent{3.19} & \EntSD{0.49} \\
        
        \NatAdv{} & -- & -- & \Ent{2.85} & \EntSD{0.22}  \\
        
        \AbsOld{} & -- & -- & \Ent{2.34} & \EntSD{0.36} \\
        
        \AbsNew{} & -- & -- & \Ent{2.33} & \EntSD{0.35} \\

        \midrule
        \textbf{TOEFL-91} (en) & -- & -- & \Ent{2.5} & \EntSD{0.39} \\

        \textbf{Hewlett} (en) & -- & -- & \Ent{2.99} & \EntSD{0.44} \\
        \bottomrule
    \end{tabularx}
    \caption{The mean and \acrfull{std} of the \entropy{} for each dataset. The last two rows display the entropy of the English datasets used by \citet{detectors_biased}: \textbf{TOEFL-91} (non-native) and \textbf{Hewlett} (native).
    }
    \label{tab:ent-mean-std}
\end{table}

\subsection{Results}

Table~\ref{tab:ent-mean-std} reveals a key finding (illustrated in Figure~\ref{fig:motivation}): non-native (\NonNative{}) speakers produce texts with greater \entropy{} than native (\NatYouth{}) speakers do (\(p<10^{-14}\)), opposed to the findings of \citet{detectors_biased}. On average, this is also the case for the advanced essays corpora (\NonNativeC{} vs. \NatAdv{}), although the difference is not significant (\(p>0.19\)).

Another finding is that the \entropy{} of the essays gets smaller as the students get more advanced (\NonNative{} vs. \NonNativeC{}; \(p < 10^{-6}\)). We investigated this on a token-level and concluded that introducing grammar errors indeed lowers the probability of the tokens of a misspelt word, increasing the \entropy{}.
Appendix~\ref{app:appendix-token-entropy} contains an illustrative example of this phenomenon.

Non-native speakers, therefore, tend to produce two types of features in the text, which have opposite effects on entropy: limited vocabulary (decreasing the entropy) and grammar errors (increasing the entropy). While \citet{detectors_biased} found that the prior is more prominent in English, we found that the latter is more prominent in Czech, which has more complex morphology.

We further observe that:
\begin{enumerate*}[label=(\arabic*)]
    \item The \entropy{} distribution does not differ significantly between \SYNmini{} and \SYNGPT{} (\(p>0.42\)).
    \item On average, \SYNllama{} has slightly smaller entropy compared to its GPT counterparts, likely caused by the reference model belonging to the same family.
    \item The \entropy{} differs across domains; e.g., both \Wiki{} and \News{} have smaller entropies compared to the SYNv9 datasets -- likely because their domains were included in pre-training data.
    \item The \entropy{} does not differ significantly between \AbsOld{} and \AbsNew{} (\(p>0.1\)).
\end{enumerate*}


\section{Generated Text Detection in Czech}\label{sec:detectors}

In order to examine question \qii{} -- whether there is a bias against non-native speakers of Czech in generated text detectors -- we work with three classes of detectors: 
\begin{enumerate*}[label=(\arabic*)]
    \item a \acrfull{nb} detector with TF-IDF features, as a baseline
    \item a fine-tuned RoBERTa-like detector, and
    \item a commercial multilingual\footnote{No functional monolingual detector for Czech exists, as of writing this article.} closed-source detector.
\end{enumerate*}

\subsection{Na\"{i}ve Bayes Detector}\label{sec:nb}

As a typical approach \citep{multi-nb}, we chose the combination of TF-IDF features and multinomial \acrshort{nb}, using the \texttt{unitok} tokeniser \citep{unitok}. We trained the detector on the \SYNtrain{} dataset with lowercasing as pre-processing and truncation to 512 tokens.
We discuss the training details in Appendix~\ref{app:nb-details}.

\newcommand{\Acc}[1]{\gradientcell{#1}{0}{100}{cyan}{yellow}{70}}

\setlength{\tabcolsep}{7pt}
\begin{table}[ht]
    \centering\footnotesize
    \begin{tabularx}{\columnwidth}{C x{0.7cm} x{0.7cm} x{0.7cm} x{0.7cm}}
        \toprule
        \textbf{Dataset} & \textbf{\acrshort{acc}} & \textbf{\acrshort{fpr}} & \textbf{\acrshort{fnr}} & \textbf{\acrshort{unk}} \\
        \midrule
        \SYNtrain{} & \Acc{99.1} & \Acc{0.8} & \Acc{1.0} & \Acc{13.1} \\
        \SYNval{} & \Acc{99.2} & \Acc{1.0} & \Acc{0.7} & \Acc{15.4} \\
        \SYNmini{} & \Acc{99.0} & -- & \Acc{1.1} & \Acc{9.4} \\
        \SYNllama{} & \Acc{73.1} & -- & \Acc{26.9} & \Acc{13.8} \\
        \Wiki{} & \Acc{88.9} & \Acc{9.4} & \Acc{13.3} & \Acc{25.1} \\
        \News{} & \Acc{98.2} & \Acc{3.4} & \Acc{0.3} & \Acc{15.0} \\
        \NonNative{} & \Acc{91.3} & \Acc{8.7} & -- & \Acc{22.2} \\
        \NonNativeC{} & \Acc{75.9} & \Acc{24.1} & -- & \Acc{14.8} \\
        \NatYouth{} & \Acc{93.8} & \Acc{6.2} & -- & \Acc{17.8} \\
        \NatAdv{} & \Acc{75.9} & \Acc{24.1} & -- & \Acc{15.8} \\
        \AbsOld{} & \Acc{19.8} & \Acc{80.2} & -- & \Acc{17.0} \\
        \AbsNew{} & \Acc{17.9} & \Acc{82.1} & -- & \Acc{17.2} \\
        \bottomrule
    \end{tabularx}
    \caption{Evaluation results of the \textbf{TF-IDF \acrshort{nb}} classifier. Values are shown as per~cent (\%). The \enquote{\acrshort{unk}} column corresponds to the ratio of out-of-vocabulary tokens in the dataset.}
    \label{tab:tf-idf-res}
\end{table}

\paragraph{Results.}
The results in Table~\ref{tab:tf-idf-res} show that \begin{enumerate*}[label=(\arabic*)]
    \item There is no significant difference in performance on the native and non-native datasets (\(p>0.23\) for \NonNative{} vs. \NatYouth{}; \(p>0.81\) for \NonNativeC{} vs. \NatAdv{}).
    \item The \acrshort{nb} detector achieves a near-perfect performance on the in-domain validation data (\SYNval{}) but deteriorates considerably on different domains.
    \item The detector is not robust to changing the source model, performing considerably worse on \SYNllama{}.
\end{enumerate*}

\subsection{RobeCzech Detector}\label{sec:rcz}

\setlength{\tabcolsep}{8pt}
\begin{table*}[ht]
    \centering\footnotesize
    \begin{tabularx}{0.85\linewidth}{ C x{0.8cm} x{0.8cm} x{0.8cm} | x{1.5cm} x{1.5cm} x{1.5cm} }
    \toprule
    \multirow{2}{*}[-0.3em]{\textbf{Dataset}} &
    \multicolumn{3}{c|}{\textbf{No Augmentation}} &
    \multicolumn{3}{c}{\textbf{Random Augmentation}} \\
    \cmidrule(lr){2-7}
    & \textbf{\acrshort{acc}} & \textbf{\acrshort{fpr}} & \textbf{\acrshort{fnr}} &
      \textbf{\acrshort{acc}} & \textbf{\acrshort{fpr}} & \textbf{\acrshort{fnr}} \\
    \midrule
    \SYNtrain{} &
    \Acc{99.4} & \Acc{0.4} & \Acc{0.9} &
    \Acc{98.9}\(\pm0.0\) & \Acc{0.7}\(\pm0.0\) & \Acc{1.4}\(\pm0.1\) \\
    \SYNval{} &
    \Acc{99.3} & \Acc{0.3} & \Acc{1.0} &
    \Acc{99.0}\(\pm0.1\) & \Acc{0.9}\(\pm0.3\) & \Acc{1.0}\(\pm0.0\) \\
    \SYNmini{} &
    \Acc{99.7} & -- & \Acc{0.4} &
    \Acc{99.7}\(\pm0.0\) & -- & \Acc{0.4}\(\pm0.0\) \\
    \SYNllama{} &
    \Acc{92.7} & -- & \Acc{7.3} &
    \Acc{92.7}\(\pm1.4\) & -- & \Acc{12.6}\(\pm1.4\) \\
    \Wiki{} &
    \Acc{86.7} & \Acc{15.2} & \Acc{11.4} &
    \Acc{86.7}\(\pm1.0\) & \Acc{10.7}\(\pm0.7\) & \Acc{15.9}\(\pm1.4\) \\
    \News{} &
    \Acc{86.0} & \Acc{27.8} & \Acc{0.3} &
    \Acc{92.7}\(\pm0.4\) & \Acc{14.0}\(\pm0.7\) & \Acc{0.6}\(\pm0.2\) \\
    \NonNative{} &
    \Acc{52.4} & \Acc{47.6} & -- &
    \Acc{67.4}\(\pm0.9\) & \Acc{32.6}\(\pm0.9\) & -- \\
    \NonNativeC{} &
    \Acc{10.3} & \Acc{89.7} & -- &
    \Acc{24.1}\(\pm2.4\) & \Acc{75.9}\(\pm2.4\) & -- \\
    \NatYouth{} &
    \Acc{71.6} & \Acc{28.4} & -- &
    \Acc{62.3}\(\pm0.7\) & \Acc{37.7}\(\pm0.7\) & -- \\
    \NatAdv{} &
    \Acc{37.9} & \Acc{62.1} & -- &
    \Acc{33.8}\(\pm2.9\) & \Acc{66.2}\(\pm2.9\) & -- \\
    \AbsOld{} &
    \Acc{33.8} & \Acc{66.2} & -- &
    \Acc{65.1}\(\pm0.5\) & \Acc{35.0}\(\pm0.5\) & -- \\
    \AbsNew{} &
    \Acc{32.4} & \Acc{67.6} & -- &
    \Acc{62.1}\(\pm0.3\) & \Acc{37.9}\(\pm0.3\) & -- \\
    \bottomrule
    \end{tabularx}
    \caption{Evaluation results of the \textbf{RobeCzech} classifier with no augmentation (left) and \acrshort{rda} (right) applied on training and inference, together with the 5-trial evaluation standard deviation (the model was only trained once). Values are shown as per~cent (\%).}
    \label{tab:rcz-res}
\end{table*}

We chose RobeCzech \citep{robeczech} as the base model for our RoBERTa-like \citep{roberta} detector, as it is the best-performing Czech monolingual model of its type. The architecture consisted of the base model and a classification head attached to the final representation of the special \texttt{[CLS]} token. The context length of RobeCzech is 512 tokens, which leads to truncation.\punctfootnote{We truncate the samples in other detectors to 512 tokens too, to provide comparable settings.} We discuss the details about the architecture and the training procedure in Appendix~\ref{app:rcz-details}.

\paragraph{Results.} The results in Table~\ref{tab:rcz-res} (left) show that, similarly to the \acrshort{nb} detector, the RobeCzech detector achieves a near-perfect performance on the training domain but lacks robustness on others. This is consistent with the findings of \citet{bert-not-robust}. Regarding the potential bias, the performance on the datasets from native speakers was considerably higher. However, upon inspection, we discovered that the relatively good performance on \NatYouth{} was caused by the presence of the \emph{non-breaking space} character, and its replacement with a regular space led to a decrease in accuracy from 71.6\% to 42.4\%, falling short of the \NonNative{} dataset.

The role of the non-breaking space is somewhat paradoxical, as it is not a part of the training dataset (or any dataset other than \NatYouth{}). We hypothesise that the model has developed some generalised rule associating rare tokens with the \emph{\enquote{human-written}} label. In an attempt to mitigate this, we introduced \acrfull{rda} using random noise. This involved adding random sequences of Unicode characters and randomly mutating the whitespace characters to sequences of other whitespace characters.
Appendix~\ref{app:rda} describes the details of the augmentation.

Table~\ref{tab:rcz-res} (right) contains the results after applying the \acrshort{rda} pipeline during training and inference. While the performance generally improved, it still did not consistently surpass the 50\% random baseline. Moreover, the detector performed inconsistently on the native vs. non-native comparison, performing better on \NonNative{} than \NatYouth{} (\(p<0.03\); \(\Delta \mathrm{FPR}=5.1\%\)) but worse on \NonNativeC{} than \NatAdv{} on average, although not significantly (\(p>0.98\)).

\subsection{Commercial Detector}\label{sec:plag}

Finally, to provide more realistic detection results, we included a commercial, closed-source model for comparison. After an informal survey of the available options, we found that \emph{Plagramme}\footnote{\url{https://www.plagramme.com/services/ai}} performs well on the tested documents. The tool operates at the sentence level, returning the classification probability for each sentence in the document. To obtain the same format as our previous detectors, we compute the average of the probabilities (each sentence with the same weight).

\setlength{\tabcolsep}{9pt}
\begin{table}[ht]
    \centering\footnotesize
    \begin{tabularx}{0.9\columnwidth}{ C c c c }
        \toprule
        \textbf{Dataset} & \textbf{\acrshort{acc}} & \textbf{\acrshort{fpr}} & \textbf{\acrshort{fnr}} \\
        \midrule
        \SYNval{} & \Acc{97.5} & \Acc{1.0} & \Acc{4.0}  \\
        
        \SYNmini{} & \Acc{99.0} & -- & \Acc{1.0}  \\
        
        \SYNllama{} & \Acc{65.0} & -- & \Acc{35.0}  \\
        
        \Wiki{} & \Acc{93.0} & \Acc{2.0} & \Acc{12.0}  \\
        
        \News{} & \Acc{96.5} & \Acc{6.0} & \Acc{1.0} \\
        
        \NonNative{} & \Acc{98.0} & \Acc{2.0} & --  \\
        
        \NonNativeC{} & \Acc{100.0} & \Acc{0.0} & --  \\
        
        \NatYouth{} & \Acc{99.0} & \Acc{1.0} & --  \\
        
        \NatAdv{} & \Acc{96.6} & \Acc{3.5} & --  \\
        
        \AbsOld{} & \Acc{96.0} & \Acc{4.0} & --  \\
        
        \AbsNew{} & \Acc{89.0} & \Acc{11.0} & --  \\
        \midrule
        \textbf{TOEFL-91} (en) & \Acc{76.9} & \Acc{23.1} & -- \\

        \textbf{Hewlett} (en) & \Acc{100.0} & \Acc{0.0} & -- \\
        \bottomrule
    \end{tabularx}
    \caption{Evaluation results of the \textbf{Plagramme} detector. Values are shown as per~cent (\%). The last two rows display the performance on the English datasets used by \citet{detectors_biased}: \textbf{TOEFL-91} (non-native) and \textbf{Hewlett} (native).}
    \label{tab:plagramme-res}
\end{table}

Due to API constraints, we limited the size of each dataset to 100 randomly selected documents, truncated each document to 512 words, and normalised the whitespace to a single space character.

\paragraph{Results.}
The results in Table~\ref{tab:plagramme-res} show considerably better performance than the classifiers we previously created and presented, indicating that our detectors fail to achieve the SoTA performance. The detector struggled the most on the \SYNllama{} dataset, suggesting that it was not trained on Llama-generated \citep{llama3} documents and does not generalise well across the models.
The difference in performance on the native vs. non-native datasets was not significant and inconsistent: performing better on \NatYouth{} than on \NonNative{} (\(p>0.11\)) but worse on \NatAdv{} than on \NonNativeC{} (\(p>0.15\)). Finally, the detector flagged 11\% of the post-GPT abstracts (\AbsNew{}) as generated, which may reflect reality.

\paragraph{Results on the English datasets.}
As a side experiment, we leveraged the detector's multilinguality to evaluate it on the datasets from \citet{detectors_biased}. While the accuracy measured on the non-native dataset was smaller than the native dataset by a non-trivial margin (\(\Delta \mathrm{FPR}=23.1\%\)), notable progress has been made since 2023:
the \acrshort{fpr} improved from the reported 61.3\% (mean) or 48\% (best detector) to our observed 23.1\%. Moreover, the correlation between text \entropy{} and detector output was negligible and slightly positive\footnote{Contrary to our expectation of a negative coefficient, as low-entropy samples supposedly receive positive labels.} (\(0 < \rho < 0.04\)), suggesting that another factor caused the drop in performance.

\subsection{Discussion}


The results presented in this section demonstrate that creating a robust detector of generated text is a feasible, yet non-trivial task. As an answer to question \qii{}, we find that \emph{none of the detectors exhibited a systematic bias against non-native speakers of Czech} when compared with their native peers. We further find that the bias against non-native speakers of English, as measured on the dataset from \citet{detectors_biased}, is considerably less pronounced in the contemporary detector than originally reported.


\section{Correlation Analysis}\label{sec:correlation}

Finally, we address question \qiii{}, whether the presented detectors rely on perplexity, or \entropy{} (in our case). While our custom detectors do not have explicit access to the \entropy{}, they may still have some internal representation of it. We test the relationship by calculating the in-class Pearson correlation coefficients between the \entropy{} (as defined in Equation~\ref{eq:our-entropy}) and the outputs of the models.

For completeness, we computed the correlation between all pairs of detectors presented in the article. \AbsNew{} was excluded to ensure that we do not compute the correlation on a potentially mixed-class dataset. We show the heatmaps for all datasets in Appendix~\ref{app:corr-maps}.

\begin{figure}
    \centering
    \includegraphics[width=0.9\columnwidth]{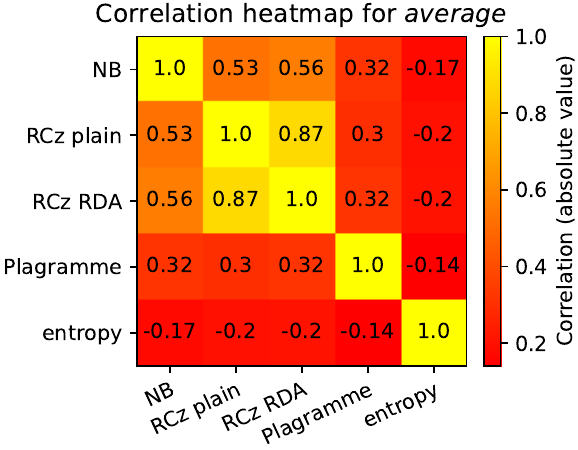}
    \caption{Per-dataset \textbf{average} correlation heatmap for the compared models. Key: NB: \acrlong{nb} detector; RCz plain: RobeCzech detector with no augmentation; RCz RDA: RobeCzech detector with \acrlong{rda}.}
    \label{fig:avg-corr}
\end{figure}

\subsection{Results}

Figure~\ref{fig:avg-corr} shows that the per-dataset mean correlation between the \entropy{} and the outputs of all models is negative, as expected (low entropy is a feature of documents with a high positive classification probability), but very weak (\(|\rho| \le 0.2\)). Moreover, the correlation between Plagramme and our custom detectors is also quite low, suggesting that they work on a different principle.

\begin{figure}[ht]
    \centering
    \includegraphics[width=0.9\columnwidth]{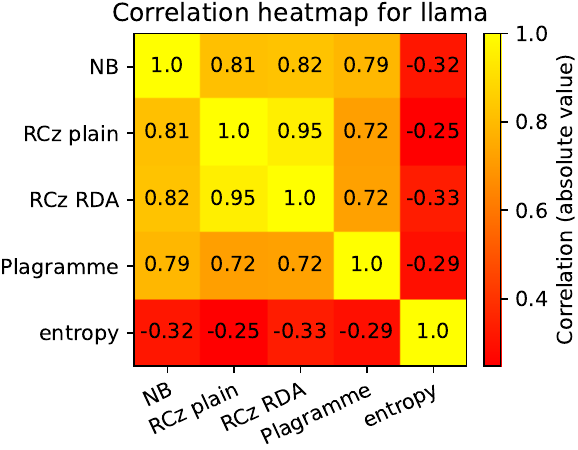}
    \caption{Correlation heatmap for the compared models, for the \SYNllama{} dataset. Key: see Figure~\ref{fig:avg-corr}.}
    \label{fig:corr-llama}
\end{figure}

Interestingly, all of the correlations were stronger in the \SYNllama{} dataset, which we show in Figure~\ref{fig:corr-llama}. This may suggest that the RobeCzech and Plagramme classifiers might work with some more-complex GPT-specific patterns but fall back to lexical features (typical for the \acrshort{nb} classifier) when those patterns are not present.

\section{Conclusion}\label{sec:conclusion}

We provide a comprehensive follow-up to the work of \citet{detectors_biased}, who claim that GPT detectors are biased against non-native speakers. Our work differs from the previous in two ways: time setting (working with the models and detectors available in 2025 rather than 2023)
and language setting (working with Czech rather than English). Under these changes, we draw considerably different conclusions. We inspect the claims using a diverse set of datasets covering, among others, essays from non-native speakers and comparable essays from native speakers.

First, we develop a method for measuring the \entropy{} of a text in a stable way. We apply this method to our datasets and find that \emph{essays by non-native speakers have \entropy{} no lower than essays by native speakers}.
In fact, it is slightly greater, which is likely due to frequent grammatical and spelling errors that contribute to entropy. The errors may be more prevalent in Czech than in English because of its complex morphology.

Next, we attempt to train our custom detectors of \acrshort{llm}-generated text from two families -- TF-IDF \acrlong{nb} and RoBERTa-like \citep{roberta} models -- and find that training a detector robust to different domains is a non-trivial task. Nonetheless, our \emph{detectors did not consistently exhibit any biases}. Moreover, we demonstrate that commercial detectors achieve satisfactory results across all domains without exhibiting bias either.

Finally, we analyse whether the presented detectors rely (explicitly or implicitly) on the \entropy{} using a correlation analysis. Our findings show that the correlation between the models' outputs and the \entropy{} is very weak (\(|\rho| \le 0.2\)), suggesting that the \emph{models do not largely depend on the \entropy{}}.

We conclude that the \emph{bias in GPT detectors is language dependent} and likely sensitive to the morphology of the specific language. Future work may conduct similar experiments on more languages to better understand the relationship. Moreover, we conclude that the technologies for detecting generated text have improved considerably since 2023, yielding more satisfactory results.

\section*{Limitations}

The sourcing of the datasets was subject to several limitations. Access to large corpora of proficient non-native speakers of Czech is limited. As a result, we used a reasonably sized dataset of essays from moderately proficient speakers (450 documents) that contained frequent errors, and a small dataset of essays from proficient speakers (29 documents). Furthermore, we only used a limited number of source \acrshort{llm}s for our documents, despite the existence of different families and more advanced models.

We encountered limitations during the creation of the detectors as well. Notably, we were unable to reach the SoTA performance with our custom detectors. We partially addressed this by including a robust, commercial detector. However, given its proprietary nature, the analysis was limited to \enquote{black-box} observations only.


\section*{Acknowledgments}

We thank Zdeněk Kasner for his valuable feedback and \emph{Plagramme} for providing access to their otherwise non-public API.

Adnan was supported by the HumanAId project \texttt{CZ.02.01.01/00/23\_025/0008691} of the Czech Ministry of Education. Jindřich H. was supported by European Union Digital Europe project no. 101195233 (OpenEuroLLM) and Horizon Europe project no. 101070350 (HPLT). Jindřich L. was supported by the CUNI project \texttt{PRIMUS/23/SCI/023} and project \texttt{CZ.02.01.01/00/23\_020/0008518} of the Czech Ministry of Education.

Computational resources were partially provided by the e-INFRA CZ project (\texttt{ID:90254}), supported by the Ministry of Education, Youth and Sports of the Czech Republic.
The work  has been using data provided by the LINDAT/CLARIAH-CZ Research Infrastructure and Czech National Corpus, supported by the Ministry of Education, Youth and Sports of the Czech Republic (projects \texttt{LM2023062} and \texttt{LM2023044}).


\bibliography{bibliography}

\appendix

\section{Dataset Generation Details}\label{app:dataset}

Listings~\ref{lst:syn-prompt}, \ref{lst:wiki-prompt}, and \ref{lst:news-prompt} show the prompts used to generate the complement for \SYNnat{}, \Wikinat{}, and \Newsnat{}, respectively.

\lstset{
  inputencoding=utf8,
  keywords={},
  breaklines=true,
  breakatwhitespace=true,
  basicstyle=\ttfamily\small, 
  literate={š}{{\v{s}}}1 {ý}{{\'{y}}}1 {í}{{\'{i}}}1 {á}{{\'{a}}}1 {é}{{\'{e}}}1 {č}{{\v{c}}}1,
  breakautoindent=false,
  breakindent=0pt,
  captionpos=b,
  frame=single,
  language={}
}

\vspace{0.5em}
\begin{lstlisting}[caption={Prompt for the synthetic text generation (only the \texttt{CS} version was used). The \texttt{<text sample>} is either the first paragraph, if its length is between 100 and 1000 characters, or the first \(n\) sentences ended by a full stop, such that \(n \in \mathbb N\) is the smallest number that makes the number of characters in the sample at least 150.}, label={lst:syn-prompt}]
[CS]
Napište dalších 2000 slov tohoto textu. Pište pouze samotný text.

<text sample>

[EN]
Write the following 2000 words of this text. Write the actual text only.

<text sample>
\end{lstlisting}

\begin{lstlisting}[caption={Prompt for the synthetic Wikipedia articles generation (only the \texttt{CS} version was used). We substituted the <article name> for the corresponding article name from the crawled articles.}, label={lst:wiki-prompt}]
[CS]
Napište Wikipedia článek na téma <article name>

[EN]
Write a Wikipedia article on the topic <article name>
\end{lstlisting}

\clearpage

\begin{lstlisting}[caption={Prompt for the synthetic news articles generation (only the \texttt{CS} version was used). We substituted the <article name> for the corresponding article name from the selected news articles. Compared to the prompt for Wikipedia, the news article names were more complex, so we introduced quotes to distinguish them from the rest of the prompt.}, label={lst:news-prompt}]
[CS]
Napište novinový článek na téma '<article name>'

[EN]
Write a news article on the topic '<article name>'
\end{lstlisting}

\section{L1 Languages of Non-Native Speakers of Czech}\label{app:l1-langs}

The distribution of native (L1) languages of the authors of \NonNative{} is the following: ru: 123 (27.3\%), zh: 70 (15.6\%), ar: 38 (8.4\%), ja: 31 (6.9\%), de: 24 (5.3\%), pl: 24 (5.3\%), en: 23 (5.1\%), fr: 21 (4.7\%), ko: 19 (4.2\%), bg: 13 (2.9\%), it: 11 (2.4\%), el: 9 (2.0\%), hu: 8 (1.8\%), fi: 6 (1.3\%), nl: 6 (1.3\%), vi: 5 (1.1\%), mo: 5 (1.1\%), uk: 4 (0.9\%), sr: 4 (0.9\%), sk: 2 (0.4\%), be: 2 (0.4\%), uz: 1 (0.2\%), no: 1 (0.2\%).

For the \NonNativeC{} dataset, the distribution is the following: ru: 17 (58.6\%), bg: 3 (10.3\%), sr: 2 (6.9\%), de: 2 (6.9\%), sk: 2 (6.9\%), ja: 2 (6.9\%), vi: 1 (3.4\%).

\section{Token-Level Entropy Analysis}\label{app:appendix-token-entropy}

\begin{figure}[H]
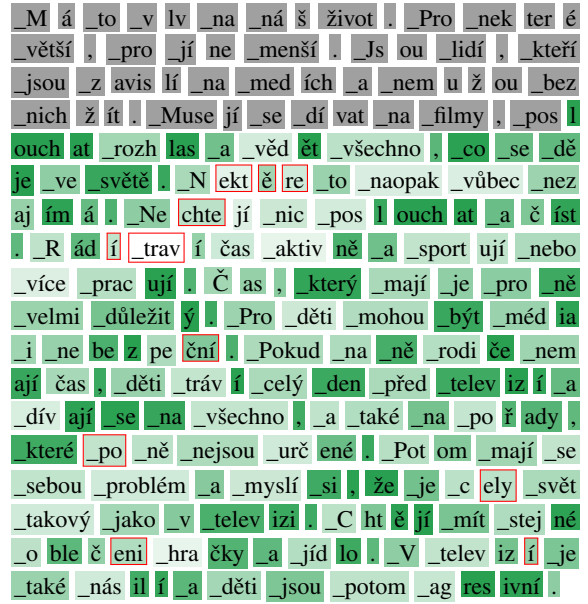

  \centering
  \begin{minipage}{0.97\linewidth}
    \footnotesize
    \fboxsep 1pt
    \colorbox[RGB]{160,160,160}{\rule{0pt}{0.75\baselineskip}\textunderscore{}M} \colorbox[RGB]{160,160,160}{\rule{0pt}{0.75\baselineskip}á} \colorbox[RGB]{160,160,160}{\rule{0pt}{0.75\baselineskip}\textunderscore{}to} \colorbox[RGB]{160,160,160}{\rule{0pt}{0.75\baselineskip}\textunderscore{}v} \colorbox[RGB]{160,160,160}{\rule{0pt}{0.75\baselineskip}lv} \colorbox[RGB]{160,160,160}{\rule{0pt}{0.75\baselineskip}\textunderscore{}na} \colorbox[RGB]{160,160,160}{\rule{0pt}{0.75\baselineskip}\textunderscore{}ná} \colorbox[RGB]{160,160,160}{\rule{0pt}{0.75\baselineskip}š} \colorbox[RGB]{160,160,160}{\rule{0pt}{0.75\baselineskip} život} \colorbox[RGB]{160,160,160}{\rule{0pt}{0.75\baselineskip}.} \colorbox[RGB]{160,160,160}{\rule{0pt}{0.75\baselineskip}\textunderscore{}Pro} \colorbox[RGB]{160,160,160}{\rule{0pt}{0.75\baselineskip}\textunderscore{}nek} \colorbox[RGB]{160,160,160}{\rule{0pt}{0.75\baselineskip}ter} \colorbox[RGB]{160,160,160}{\rule{0pt}{0.75\baselineskip}é} \colorbox[RGB]{160,160,160}{\rule{0pt}{0.75\baselineskip}\textunderscore{}větší} \colorbox[RGB]{160,160,160}{\rule{0pt}{0.75\baselineskip},} \colorbox[RGB]{160,160,160}{\rule{0pt}{0.75\baselineskip}\textunderscore{}pro} \colorbox[RGB]{160,160,160}{\rule{0pt}{0.75\baselineskip}\textunderscore{}jí} \colorbox[RGB]{160,160,160}{\rule{0pt}{0.75\baselineskip}ne} \colorbox[RGB]{160,160,160}{\rule{0pt}{0.75\baselineskip}\textunderscore{}menší} \colorbox[RGB]{160,160,160}{\rule{0pt}{0.75\baselineskip}.} \colorbox[RGB]{160,160,160}{\rule{0pt}{0.75\baselineskip}\textunderscore{}Js} \colorbox[RGB]{160,160,160}{\rule{0pt}{0.75\baselineskip}ou} \colorbox[RGB]{160,160,160}{\rule{0pt}{0.75\baselineskip}\textunderscore{}lidí} \colorbox[RGB]{160,160,160}{\rule{0pt}{0.75\baselineskip},} \colorbox[RGB]{160,160,160}{\rule{0pt}{0.75\baselineskip}\textunderscore{}kteří} \colorbox[RGB]{160,160,160}{\rule{0pt}{0.75\baselineskip}\textunderscore{}jsou} \colorbox[RGB]{160,160,160}{\rule{0pt}{0.75\baselineskip}\textunderscore{}z} \colorbox[RGB]{160,160,160}{\rule{0pt}{0.75\baselineskip}avis} \colorbox[RGB]{160,160,160}{\rule{0pt}{0.75\baselineskip}lí} \colorbox[RGB]{160,160,160}{\rule{0pt}{0.75\baselineskip}\textunderscore{}na} \colorbox[RGB]{160,160,160}{\rule{0pt}{0.75\baselineskip}\textunderscore{}med} \colorbox[RGB]{160,160,160}{\rule{0pt}{0.75\baselineskip}ích} \colorbox[RGB]{160,160,160}{\rule{0pt}{0.75\baselineskip}\textunderscore{}a} \colorbox[RGB]{160,160,160}{\rule{0pt}{0.75\baselineskip}\textunderscore{}nem} \colorbox[RGB]{160,160,160}{\rule{0pt}{0.75\baselineskip}u} \colorbox[RGB]{160,160,160}{\rule{0pt}{0.75\baselineskip}ž} \colorbox[RGB]{160,160,160}{\rule{0pt}{0.75\baselineskip}ou} \colorbox[RGB]{160,160,160}{\rule{0pt}{0.75\baselineskip}\textunderscore{}bez} \colorbox[RGB]{160,160,160}{\rule{0pt}{0.75\baselineskip}\textunderscore{}nich} \colorbox[RGB]{160,160,160}{\rule{0pt}{0.75\baselineskip} ž} \colorbox[RGB]{160,160,160}{\rule{0pt}{0.75\baselineskip}ít} \colorbox[RGB]{160,160,160}{\rule{0pt}{0.75\baselineskip}.} \colorbox[RGB]{160,160,160}{\rule{0pt}{0.75\baselineskip}\textunderscore{}Muse} \colorbox[RGB]{160,160,160}{\rule{0pt}{0.75\baselineskip}jí} \colorbox[RGB]{160,160,160}{\rule{0pt}{0.75\baselineskip}\textunderscore{}se} \colorbox[RGB]{160,160,160}{\rule{0pt}{0.75\baselineskip}\textunderscore{}dí} \colorbox[RGB]{160,160,160}{\rule{0pt}{0.75\baselineskip}vat} \colorbox[RGB]{160,160,160}{\rule{0pt}{0.75\baselineskip}\textunderscore{}na} \colorbox[RGB]{160,160,160}{\rule{0pt}{0.75\baselineskip}\textunderscore{}filmy} \colorbox[RGB]{160,160,160}{\rule{0pt}{0.75\baselineskip},} \colorbox[RGB]{160,160,160}{\rule{0pt}{0.75\baselineskip}\textunderscore{}pos} \colorbox[RGB]{42,161,81}{\rule{0pt}{0.75\baselineskip}l} \colorbox[RGB]{41,160,81}{\rule{0pt}{0.75\baselineskip}ouch} \colorbox[RGB]{41,160,81}{\rule{0pt}{0.75\baselineskip}at} \colorbox[RGB]{163,215,180}{\rule{0pt}{0.75\baselineskip}\textunderscore{}rozh} \colorbox[RGB]{42,161,82}{\rule{0pt}{0.75\baselineskip}las} \colorbox[RGB]{114,193,140}{\rule{0pt}{0.75\baselineskip}\textunderscore{}a} \colorbox[RGB]{210,235,218}{\rule{0pt}{0.75\baselineskip}\textunderscore{}věd} \colorbox[RGB]{40,160,80}{\rule{0pt}{0.75\baselineskip}ět} \colorbox[RGB]{188,225,201}{\rule{0pt}{0.75\baselineskip}\textunderscore{}všechno} \colorbox[RGB]{90,182,121}{\rule{0pt}{0.75\baselineskip},} \colorbox[RGB]{43,161,83}{\rule{0pt}{0.75\baselineskip}\textunderscore{}co} \colorbox[RGB]{104,188,132}{\rule{0pt}{0.75\baselineskip}\textunderscore{}se} \colorbox[RGB]{97,185,127}{\rule{0pt}{0.75\baselineskip}\textunderscore{}dě} \colorbox[RGB]{41,160,81}{\rule{0pt}{0.75\baselineskip}je} \colorbox[RGB]{143,205,164}{\rule{0pt}{0.75\baselineskip}\textunderscore{}ve} \colorbox[RGB]{56,167,93}{\rule{0pt}{0.75\baselineskip}\textunderscore{}světě} \colorbox[RGB]{57,167,94}{\rule{0pt}{0.75\baselineskip}.} \colorbox[RGB]{169,217,185}{\rule{0pt}{0.75\baselineskip}\textunderscore{}N} \fcolorbox[RGB]{255,0,0}{228,243,233}{\rule{0pt}{0.75\baselineskip}ekt} \fcolorbox[RGB]{255,0,0}{162,214,179}{\rule{0pt}{0.75\baselineskip}ě} \fcolorbox[RGB]{255,0,0}{207,234,216}{\rule{0pt}{0.75\baselineskip}re} \colorbox[RGB]{153,210,172}{\rule{0pt}{0.75\baselineskip}\textunderscore{}to} \colorbox[RGB]{217,238,224}{\rule{0pt}{0.75\baselineskip}\textunderscore{}naopak} \colorbox[RGB]{218,238,225}{\rule{0pt}{0.75\baselineskip}\textunderscore{}vůbec} \colorbox[RGB]{157,212,175}{\rule{0pt}{0.75\baselineskip}\textunderscore{}nez} \colorbox[RGB]{151,209,170}{\rule{0pt}{0.75\baselineskip}aj} \colorbox[RGB]{46,162,85}{\rule{0pt}{0.75\baselineskip}ím} \colorbox[RGB]{101,187,130}{\rule{0pt}{0.75\baselineskip}á} \colorbox[RGB]{78,177,111}{\rule{0pt}{0.75\baselineskip}.} \colorbox[RGB]{142,205,163}{\rule{0pt}{0.75\baselineskip}\textunderscore{}Ne} \fcolorbox[RGB]{255,0,0}{188,225,201}{\rule{0pt}{0.75\baselineskip}chte} \colorbox[RGB]{222,240,228}{\rule{0pt}{0.75\baselineskip}jí} \colorbox[RGB]{204,232,213}{\rule{0pt}{0.75\baselineskip}\textunderscore{}nic} \colorbox[RGB]{210,235,218}{\rule{0pt}{0.75\baselineskip}\textunderscore{}pos} \colorbox[RGB]{57,168,94}{\rule{0pt}{0.75\baselineskip}l} \colorbox[RGB]{42,161,82}{\rule{0pt}{0.75\baselineskip}ouch} \colorbox[RGB]{40,160,80}{\rule{0pt}{0.75\baselineskip}at} \colorbox[RGB]{106,189,134}{\rule{0pt}{0.75\baselineskip}\textunderscore{}a} \colorbox[RGB]{128,199,151}{\rule{0pt}{0.75\baselineskip} č} \colorbox[RGB]{49,164,87}{\rule{0pt}{0.75\baselineskip}íst} \colorbox[RGB]{95,184,125}{\rule{0pt}{0.75\baselineskip}.} \colorbox[RGB]{190,226,202}{\rule{0pt}{0.75\baselineskip}\textunderscore{}R} \colorbox[RGB]{94,184,124}{\rule{0pt}{0.75\baselineskip}ád} \fcolorbox[RGB]{255,0,0}{193,228,204}{\rule{0pt}{0.75\baselineskip}í} \fcolorbox[RGB]{255,0,0}{255,255,255}{\rule{0pt}{0.75\baselineskip}\textunderscore{}trav} \colorbox[RGB]{148,208,168}{\rule{0pt}{0.75\baselineskip}í} \colorbox[RGB]{205,233,215}{\rule{0pt}{0.75\baselineskip} čas} \colorbox[RGB]{230,244,234}{\rule{0pt}{0.75\baselineskip}\textunderscore{}aktiv} \colorbox[RGB]{75,176,109}{\rule{0pt}{0.75\baselineskip}ně} \colorbox[RGB]{112,192,139}{\rule{0pt}{0.75\baselineskip}\textunderscore{}a} \colorbox[RGB]{194,228,205}{\rule{0pt}{0.75\baselineskip}\textunderscore{}sport} \colorbox[RGB]{191,227,203}{\rule{0pt}{0.75\baselineskip}ují} \colorbox[RGB]{174,219,189}{\rule{0pt}{0.75\baselineskip}\textunderscore{}nebo} \colorbox[RGB]{220,239,226}{\rule{0pt}{0.75\baselineskip}\textunderscore{}více} \colorbox[RGB]{182,223,196}{\rule{0pt}{0.75\baselineskip}\textunderscore{}prac} \colorbox[RGB]{41,161,81}{\rule{0pt}{0.75\baselineskip}ují} \colorbox[RGB]{82,179,114}{\rule{0pt}{0.75\baselineskip}.} \colorbox[RGB]{163,215,180}{\rule{0pt}{0.75\baselineskip} Č} \colorbox[RGB]{149,208,168}{\rule{0pt}{0.75\baselineskip}as} \colorbox[RGB]{139,204,160}{\rule{0pt}{0.75\baselineskip},} \colorbox[RGB]{50,165,88}{\rule{0pt}{0.75\baselineskip}\textunderscore{}který} \colorbox[RGB]{169,217,185}{\rule{0pt}{0.75\baselineskip}\textunderscore{}mají} \colorbox[RGB]{144,206,165}{\rule{0pt}{0.75\baselineskip}\textunderscore{}je} \colorbox[RGB]{140,204,162}{\rule{0pt}{0.75\baselineskip}\textunderscore{}pro} \colorbox[RGB]{61,169,97}{\rule{0pt}{0.75\baselineskip}\textunderscore{}ně} \colorbox[RGB]{155,211,173}{\rule{0pt}{0.75\baselineskip}\textunderscore{}velmi} \colorbox[RGB]{119,195,144}{\rule{0pt}{0.75\baselineskip}\textunderscore{}důležit} \colorbox[RGB]{40,160,80}{\rule{0pt}{0.75\baselineskip}ý} \colorbox[RGB]{61,169,97}{\rule{0pt}{0.75\baselineskip}.} \colorbox[RGB]{157,212,175}{\rule{0pt}{0.75\baselineskip}\textunderscore{}Pro} \colorbox[RGB]{205,233,214}{\rule{0pt}{0.75\baselineskip}\textunderscore{}děti} \colorbox[RGB]{177,220,191}{\rule{0pt}{0.75\baselineskip}\textunderscore{}mohou} \colorbox[RGB]{57,167,94}{\rule{0pt}{0.75\baselineskip}\textunderscore{}být} \colorbox[RGB]{153,210,172}{\rule{0pt}{0.75\baselineskip}\textunderscore{}méd} \colorbox[RGB]{46,163,85}{\rule{0pt}{0.75\baselineskip}ia} \colorbox[RGB]{159,213,177}{\rule{0pt}{0.75\baselineskip}\textunderscore{}i} \colorbox[RGB]{151,209,171}{\rule{0pt}{0.75\baselineskip}\textunderscore{}ne} \colorbox[RGB]{76,176,110}{\rule{0pt}{0.75\baselineskip}be} \colorbox[RGB]{40,160,80}{\rule{0pt}{0.75\baselineskip}z} \colorbox[RGB]{174,219,189}{\rule{0pt}{0.75\baselineskip}pe} \fcolorbox[RGB]{255,0,0}{131,200,154}{\rule{0pt}{0.75\baselineskip}ční} \colorbox[RGB]{77,176,110}{\rule{0pt}{0.75\baselineskip}.} \colorbox[RGB]{172,218,187}{\rule{0pt}{0.75\baselineskip}\textunderscore{}Pokud} \colorbox[RGB]{174,219,189}{\rule{0pt}{0.75\baselineskip}\textunderscore{}na} \colorbox[RGB]{71,174,105}{\rule{0pt}{0.75\baselineskip}\textunderscore{}ně} \colorbox[RGB]{181,223,195}{\rule{0pt}{0.75\baselineskip}\textunderscore{}rodi} \colorbox[RGB]{42,161,81}{\rule{0pt}{0.75\baselineskip}če} \colorbox[RGB]{158,212,176}{\rule{0pt}{0.75\baselineskip}\textunderscore{}nem} \colorbox[RGB]{56,167,93}{\rule{0pt}{0.75\baselineskip}ají} \colorbox[RGB]{167,216,183}{\rule{0pt}{0.75\baselineskip} čas} \colorbox[RGB]{60,169,96}{\rule{0pt}{0.75\baselineskip},} \colorbox[RGB]{152,209,171}{\rule{0pt}{0.75\baselineskip}\textunderscore{}děti} \colorbox[RGB]{195,229,206}{\rule{0pt}{0.75\baselineskip}\textunderscore{}tráv} \colorbox[RGB]{42,161,81}{\rule{0pt}{0.75\baselineskip}í} \colorbox[RGB]{171,218,186}{\rule{0pt}{0.75\baselineskip}\textunderscore{}celý} \colorbox[RGB]{60,169,96}{\rule{0pt}{0.75\baselineskip}\textunderscore{}den} \colorbox[RGB]{171,218,187}{\rule{0pt}{0.75\baselineskip}\textunderscore{}před} \colorbox[RGB]{62,170,98}{\rule{0pt}{0.75\baselineskip}\textunderscore{}telev} \colorbox[RGB]{41,160,80}{\rule{0pt}{0.75\baselineskip}iz} \colorbox[RGB]{51,165,89}{\rule{0pt}{0.75\baselineskip}í} \colorbox[RGB]{100,186,128}{\rule{0pt}{0.75\baselineskip}\textunderscore{}a} \colorbox[RGB]{216,238,223}{\rule{0pt}{0.75\baselineskip}\textunderscore{}dív} \colorbox[RGB]{45,162,84}{\rule{0pt}{0.75\baselineskip}ají} \colorbox[RGB]{43,161,82}{\rule{0pt}{0.75\baselineskip}\textunderscore{}se} \colorbox[RGB]{55,166,92}{\rule{0pt}{0.75\baselineskip}\textunderscore{}na} \colorbox[RGB]{195,229,206}{\rule{0pt}{0.75\baselineskip}\textunderscore{}všechno} \colorbox[RGB]{66,172,101}{\rule{0pt}{0.75\baselineskip},} \colorbox[RGB]{191,227,203}{\rule{0pt}{0.75\baselineskip}\textunderscore{}a} \colorbox[RGB]{177,221,192}{\rule{0pt}{0.75\baselineskip}\textunderscore{}také} \colorbox[RGB]{118,194,143}{\rule{0pt}{0.75\baselineskip}\textunderscore{}na} \colorbox[RGB]{196,229,207}{\rule{0pt}{0.75\baselineskip}\textunderscore{}po} \colorbox[RGB]{83,179,115}{\rule{0pt}{0.75\baselineskip}ř} \colorbox[RGB]{85,180,117}{\rule{0pt}{0.75\baselineskip}ady} \colorbox[RGB]{97,185,126}{\rule{0pt}{0.75\baselineskip},} \colorbox[RGB]{62,170,98}{\rule{0pt}{0.75\baselineskip}\textunderscore{}které} \fcolorbox[RGB]{255,0,0}{194,228,205}{\rule{0pt}{0.75\baselineskip}\textunderscore{}po} \colorbox[RGB]{187,225,200}{\rule{0pt}{0.75\baselineskip}\textunderscore{}ně} \colorbox[RGB]{173,219,188}{\rule{0pt}{0.75\baselineskip}\textunderscore{}nejsou} \colorbox[RGB]{207,234,216}{\rule{0pt}{0.75\baselineskip}\textunderscore{}urč} \colorbox[RGB]{106,189,134}{\rule{0pt}{0.75\baselineskip}ené} \colorbox[RGB]{58,168,95}{\rule{0pt}{0.75\baselineskip}.} \colorbox[RGB]{182,223,196}{\rule{0pt}{0.75\baselineskip}\textunderscore{}Pot} \colorbox[RGB]{133,201,156}{\rule{0pt}{0.75\baselineskip}om} \colorbox[RGB]{162,214,179}{\rule{0pt}{0.75\baselineskip}\textunderscore{}mají} \colorbox[RGB]{194,228,205}{\rule{0pt}{0.75\baselineskip}\textunderscore{}se} \colorbox[RGB]{193,228,204}{\rule{0pt}{0.75\baselineskip}\textunderscore{}sebou} \colorbox[RGB]{207,234,216}{\rule{0pt}{0.75\baselineskip}\textunderscore{}problém} \colorbox[RGB]{136,202,158}{\rule{0pt}{0.75\baselineskip}\textunderscore{}a} \colorbox[RGB]{204,232,213}{\rule{0pt}{0.75\baselineskip}\textunderscore{}myslí} \colorbox[RGB]{63,170,99}{\rule{0pt}{0.75\baselineskip}\textunderscore{}si} \colorbox[RGB]{43,161,82}{\rule{0pt}{0.75\baselineskip},} \colorbox[RGB]{41,160,81}{\rule{0pt}{0.75\baselineskip} že} \colorbox[RGB]{115,193,141}{\rule{0pt}{0.75\baselineskip}\textunderscore{}je} \colorbox[RGB]{188,225,201}{\rule{0pt}{0.75\baselineskip}\textunderscore{}c} \fcolorbox[RGB]{255,0,0}{189,226,202}{\rule{0pt}{0.75\baselineskip}ely} \colorbox[RGB]{171,218,187}{\rule{0pt}{0.75\baselineskip}\textunderscore{}svět} \colorbox[RGB]{207,234,216}{\rule{0pt}{0.75\baselineskip}\textunderscore{}takový} \colorbox[RGB]{164,215,181}{\rule{0pt}{0.75\baselineskip}\textunderscore{}jako} \colorbox[RGB]{134,201,156}{\rule{0pt}{0.75\baselineskip}\textunderscore{}v} \colorbox[RGB]{76,176,109}{\rule{0pt}{0.75\baselineskip}\textunderscore{}telev} \colorbox[RGB]{59,169,96}{\rule{0pt}{0.75\baselineskip}izi} \colorbox[RGB]{64,171,100}{\rule{0pt}{0.75\baselineskip}.} \colorbox[RGB]{187,225,199}{\rule{0pt}{0.75\baselineskip}\textunderscore{}C} \colorbox[RGB]{128,199,151}{\rule{0pt}{0.75\baselineskip}ht} \colorbox[RGB]{45,162,84}{\rule{0pt}{0.75\baselineskip}ě} \colorbox[RGB]{40,160,80}{\rule{0pt}{0.75\baselineskip}jí} \colorbox[RGB]{122,196,147}{\rule{0pt}{0.75\baselineskip}\textunderscore{}mít} \colorbox[RGB]{175,220,190}{\rule{0pt}{0.75\baselineskip}\textunderscore{}stej} \colorbox[RGB]{45,162,84}{\rule{0pt}{0.75\baselineskip}né} \colorbox[RGB]{188,225,200}{\rule{0pt}{0.75\baselineskip}\textunderscore{}o} \colorbox[RGB]{93,184,123}{\rule{0pt}{0.75\baselineskip}ble} \colorbox[RGB]{147,207,167}{\rule{0pt}{0.75\baselineskip}č} \fcolorbox[RGB]{255,0,0}{176,220,191}{\rule{0pt}{0.75\baselineskip}eni} \colorbox[RGB]{232,245,236}{\rule{0pt}{0.75\baselineskip}\textunderscore{}hra} \colorbox[RGB]{82,179,114}{\rule{0pt}{0.75\baselineskip}čky} \colorbox[RGB]{95,184,125}{\rule{0pt}{0.75\baselineskip}\textunderscore{}a} \colorbox[RGB]{180,222,194}{\rule{0pt}{0.75\baselineskip}\textunderscore{}jíd} \colorbox[RGB]{61,169,97}{\rule{0pt}{0.75\baselineskip}lo} \colorbox[RGB]{94,184,124}{\rule{0pt}{0.75\baselineskip}.} \colorbox[RGB]{143,205,164}{\rule{0pt}{0.75\baselineskip}\textunderscore{}V} \colorbox[RGB]{194,228,205}{\rule{0pt}{0.75\baselineskip}\textunderscore{}telev} \colorbox[RGB]{118,194,143}{\rule{0pt}{0.75\baselineskip}iz} \fcolorbox[RGB]{255,0,0}{149,208,168}{\rule{0pt}{0.75\baselineskip}í} \colorbox[RGB]{124,197,149}{\rule{0pt}{0.75\baselineskip}\textunderscore{}je} \colorbox[RGB]{153,210,172}{\rule{0pt}{0.75\baselineskip}\textunderscore{}také} \colorbox[RGB]{207,234,216}{\rule{0pt}{0.75\baselineskip}\textunderscore{}nás} \colorbox[RGB]{66,171,101}{\rule{0pt}{0.75\baselineskip}il} \colorbox[RGB]{52,165,89}{\rule{0pt}{0.75\baselineskip}í} \colorbox[RGB]{92,183,122}{\rule{0pt}{0.75\baselineskip}\textunderscore{}a} \colorbox[RGB]{164,215,181}{\rule{0pt}{0.75\baselineskip}\textunderscore{}děti} \colorbox[RGB]{126,198,150}{\rule{0pt}{0.75\baselineskip}\textunderscore{}jsou} \colorbox[RGB]{190,226,202}{\rule{0pt}{0.75\baselineskip}\textunderscore{}potom} \colorbox[RGB]{200,231,210}{\rule{0pt}{0.75\baselineskip}\textunderscore{}ag} \colorbox[RGB]{40,160,80}{\rule{0pt}{0.75\baselineskip}res} \colorbox[RGB]{47,163,86}{\rule{0pt}{0.75\baselineskip}ivní} \colorbox[RGB]{142,205,163}{\rule{0pt}{0.75\baselineskip}.}
  \end{minipage}
  \caption{Tokens' contributions to entropy, written by a non-native speaker. Darker tokens have higher likelihoods. Start-word tokens begin with an underscore (\textunderscore{}). The first 50 tokens (in \colorbox[RGB]{160,160,160}{grey}) are used to introduce the context, and their likelihood is not measured. Non-introductory tokens with spelling or grammatical errors have \fcolorbox[RGB]{255,0,0}{255,255,255}{red borders}.}
  \label{fig:entropy-contrib}
\end{figure}

We analysed the entropy of selected texts from non-native speakers qualitatively to understand their increased entropy, and concluded that grammar errors have a prominent role in this phenomenon. We show an illustrative example in Figure~\ref{fig:entropy-contrib}. Notice that for most correct words split into multiple tokens, the first token is often difficult to predict, yet the remaining tokens are quite predictable from the context. This, however, does \emph{not} hold for misspelt words, in which even the later tokens have a low probability.

\section{Na\"{i}ve Bayes Details}\label{app:nb-details}

In this section, we discuss the training details of the TF-IDF Na\"{i}ve Bayes detector. The detextor was trained on the \SYNtrain{} dataset. For text vectorisation, we used the \texttt{TfidfVectorizer} from the \texttt{scikit-learn} library \citep{sklearn}. For the \acrshort{nb} implementation, we used the \texttt{MultinomialNB} from the same library. Other than the \texttt{unitok} tokeniser, we used the default parameters. The vocabulary size (feature vector dimension) was 162\,109.

We further experimented with using the RobeCzech tokeniser instead of \texttt{unitok} and got comparable results, shown in Table~\ref{tab:tf-idf-rcz-res}. The vocabulary size was 49\,998.

\begin{table}[ht]
    \centering\footnotesize
    \setlength{\tabcolsep}{6.5pt}
    \begin{tabularx}{\columnwidth}{C x{0.7cm} x{0.7cm} x{0.7cm} x{0.7cm}}
        \toprule
        \textbf{Dataset} & \textbf{\acrshort{acc}} & \textbf{\acrshort{fpr}} & \textbf{\acrshort{fnr}} & \textbf{\acrshort{unk}} \\
        \midrule
        \SYNtrain{} & \Acc{98.7} & \Acc{0.6} & \Acc{2.1} & \Acc{0.0} \\
        
        \SYNval{} & \Acc{98.8} & \Acc{0.3} & \Acc{2.0} & \Acc{0.1}  \\
        
        \SYNmini{} & \Acc{99.0} & -- & \Acc{1.0} & \Acc{0.0}  \\
        
        \SYNllama{} & \Acc{72.4} & -- & \Acc{27.6} & \Acc{0.1}  \\
        
        \Wiki{} & \Acc{80.3} & \Acc{4.9} & \Acc{39.8} & \Acc{0.5}  \\
        
        \News{} & \Acc{96.7} & \Acc{6.3} & \Acc{0.3} & \Acc{0.2}  \\
        
        \NonNative{} & \Acc{93.1} & \Acc{6.9} & -- & \Acc{0.3}  \\
        
        \NonNativeC{} & \Acc{65.5} & \Acc{34.5} & -- & \Acc{0.2}  \\
        
        \NatYouth{} & \Acc{93.6} & \Acc{6.4} & -- & \Acc{0.2}  \\
        
        \NatAdv{} & \Acc{79.3} & \Acc{20.7} & -- & \Acc{0.2}  \\
        
        \AbsOld{} & \Acc{39.5} & \Acc{60.5} & -- & \Acc{0.2}  \\
        
        \AbsNew{} & \Acc{38.5} & \Acc{61.5} & -- & \Acc{0.2}  \\
        \bottomrule
    \end{tabularx}
    \caption{Evaluation results of the TF-IDF \acrshort{nb} classifier with the RobeCzech tokeniser. Values are shown as per~cent (\%). The \enquote{UnkR} column corresponds to the ratio of out-of-vocabulary tokens in the dataset (also in per~cent).}
    \label{tab:tf-idf-rcz-res}
\end{table}

\section{RobeCzech Details}\label{app:rcz-details}

In this section, we describe in detail the architecture, training procedure and data augmentation of the RobeCzech classifier.

\subsection{Architecture}

\begin{figure}[h]
    \centering
    \includegraphics[width=\columnwidth]{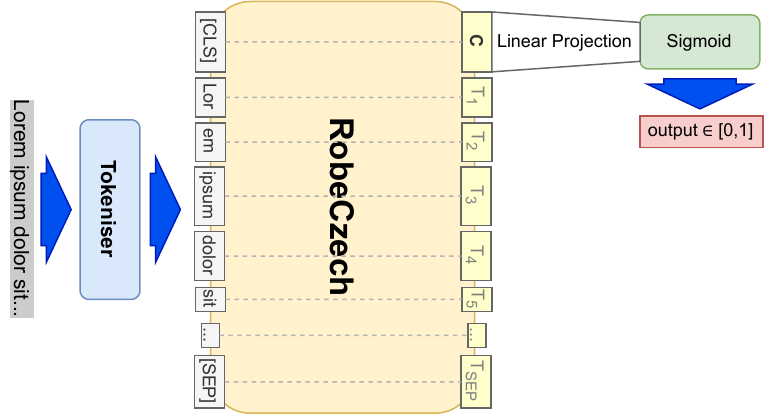}
    \caption{The architecture of the RobeCzech classifier.}
    \label{fig:robeczech-arch}
\end{figure}

The architecture is based on the architecture for sentiment analysis described by \citet[Sec. 4.6]{robeczech}. The prediction works as follows:
\begin{enumerate}
    \item The input text is tokenised, and special tokens are added; importantly, the \texttt{[CLS]} token at the beginning.
    \item The tokens are passed through RobeCzech, and the output of the last hidden layer (i.e. the contextualised embeddings) is extracted.
    \item The embedding of the \texttt{[CLS]} token is linearly projected to dimension 1. The rest of the embeddings are disregarded.
    \item The linear projection is followed by a sigmoid activation, resulting in output \(\in [0,1]\) -- the probability of positive classification.
\end{enumerate}
The architecture is illustrated in Figure \ref{fig:robeczech-arch}.

\subsection{Training}

We trained the classifier on the \SYNtrain{} dataset with no text normalisation. We used the \SYNval{} dataset for hyperparameter tuning. We trained the model on a single NVIDIA RTX A4000 GPU (16~GB VRAM). We used the PyTorch \citep{pytorch} implementation of the standard training procedures.

The training procedure consisted of three phases: in the first phase, we froze the RobeCzech parameters and only trained the linear projection layer (classification head) at a constant learning rate. In the second phase, we unfroze the RobeCzech weights and trained them with a linear learning rate warmup from 0 to a specified value. Finally, in the third phase, we kept the weights unfrozen and trained with cosine decay to 0.

We used the following hyperparameters in all the phases: batch size: \(32\), optimiser: AdamW (with the default \(\beta_1=0.9\), \(\beta_2=0.999\)), weight decay: \(10^{-3}\), label smoothing: \(0.1\).

The classification head was trained by itself for one epoch at a learning rate \(5\times 10^{-4}\) (\(10^{-3}\) with \acrshort{rda}). We were able to reach the accuracy of 89.80\% on \SYNval{} after this epoch alone. Next, we trained the whole model with a linear learning rate warmup from 0 to \(3\times 10^{-7}\) over one epoch. Finally, the model was trained for an additional three epochs (1 epoch with \acrshort{rda}) with cosine learning rate decay to \(0\).

\subsection{Random Augmentation}\label{app:rda}

In order to make the classifier more robust against rare characters, we introduced \acrfull{rda}. The process first employs the \texttt{sacremoses}\footnote{\url{https://github.com/hplt-project/sacremoses}} punctuation normaliser and then adds random Unicode and whitespace noise.

\newcommand{\round}[1]{\ensuremath{\lfloor#1\rceil}}
Adding random Unicode noise involves inserting random \enquote{words} -- sequences of randomly generated printable Unicode symbols. First, the number of words to add is determined: for a sequence of \(w\) words and the expected inflation factor of \(0.02\), the number of added words will be \(\max\left(0, \round{x}\right)\), where \(x \sim \mathcal{N}\left(0.02w, \frac{0.02w}{5}\right)\). Next, each word is generated by determining its length as \(\max(0, \round{y})\), where \(y \sim \mathcal{N}(1,1)\), and generating such a sequence of characters. The words are then inserted into positions generated at random, with repetition.

After adding random words, we join both the original and the inserted words with random whitespace. With the probability of 0.97, we use a single space character. Otherwise, we use a sequence of \(\max(1, \round{z})\) where \(z \sim \mathcal{N}(1,0.2)\), whitespace characters from the following list: \texttt{\textbackslash n}, \texttt{\textbackslash t}, \texttt{\textbackslash r\textbackslash n}, \texttt{\textbackslash n\textbackslash n}, \texttt{\textbackslash r\textbackslash n\textbackslash r\textbackslash n}, and \texttt{\textbackslash u00a0} (the \emph{non-breaking space} -- \texttt{\&nbsp}).

\clearpage
\onecolumn
\section{Entropy Distributions}\label{app:entropy-dist}
\begin{figure}[H]
    \centering
    \includegraphics[width=0.95\linewidth]{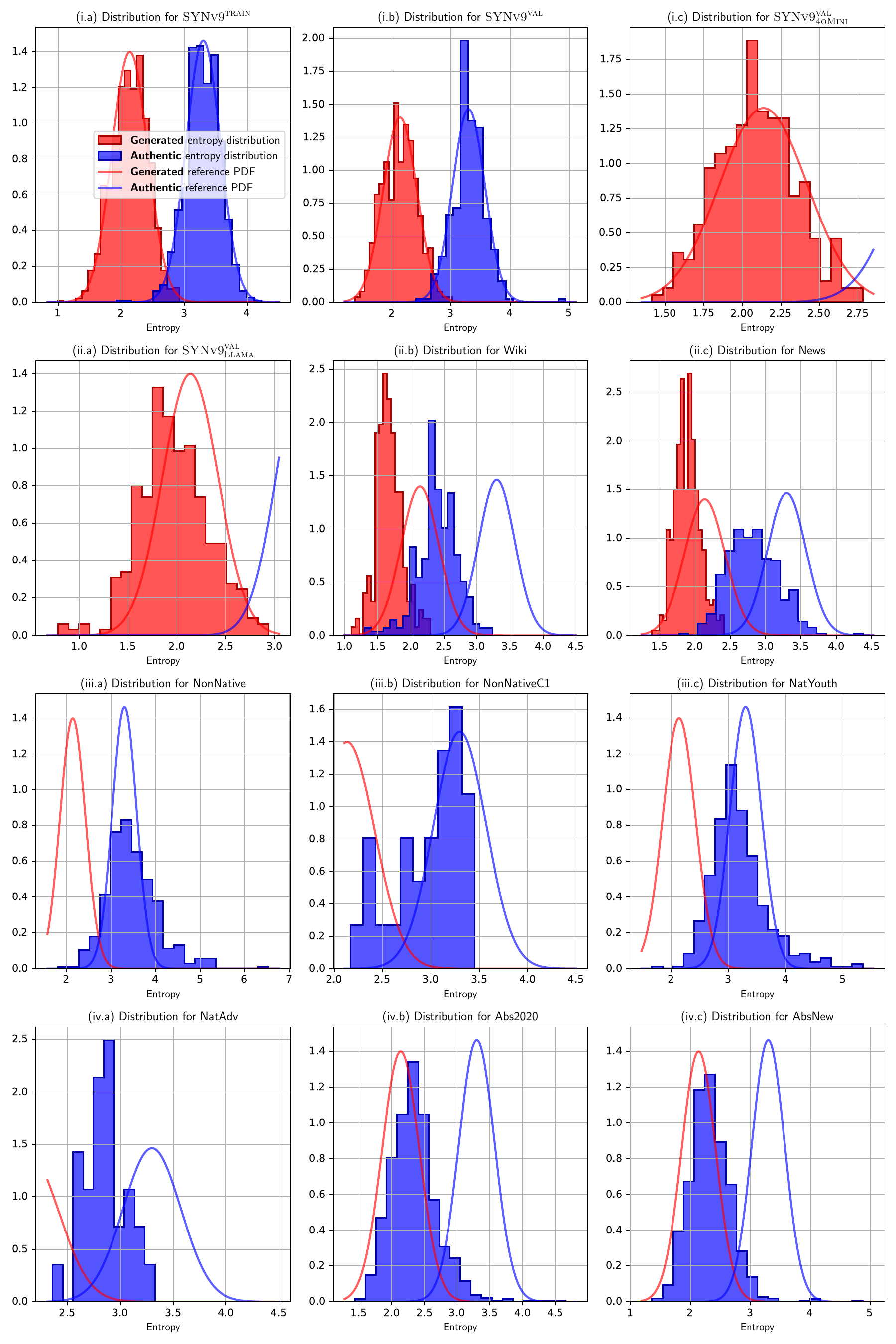}
    \caption{Entropy distribution densities of all the datasets compared to the density fitted on \SYNtrain{}.}
    \label{fig:all_entr}
\end{figure}


\section{Correlation Heatmaps}\label{app:corr-maps}
\begin{figure}[H]
    \centering
    \includegraphics[width=0.99\linewidth]{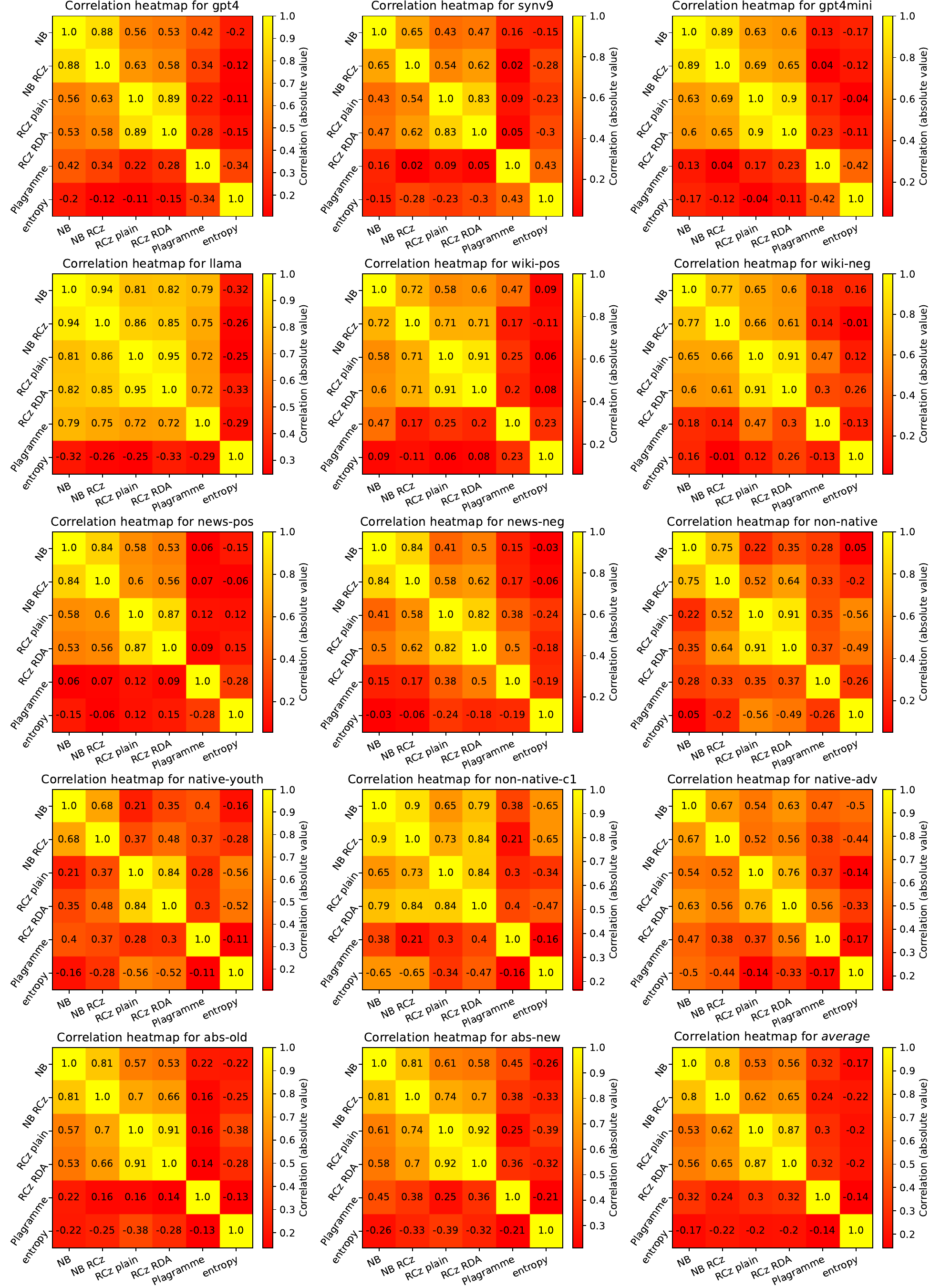}
    \caption{Correlation heatmap for the compared models, for each dataset. Key: NB: \acrlong{nb} detector; NB RCz: \acrlong{nb} detector with the RobeCzech tokeniser; RCz plain: RobeCzech detector with no augmentation, RCz RDA: RobeCzech detector with \acrlong{rda}.}
    \label{fig:placeholder}
\end{figure}

\end{document}